\documentclass[10pt,logo,copyright]{nvidiatechreport}
\usepackage[authoryear,sort&compress,round]{natbib}

\usepackage[utf8]{inputenc} % allow utf-8 input
\usepackage[T1]{fontenc}    % use 8-bit T1 fonts

\usepackage{parskip}        % no paragraph indents
\usepackage{url}            % simple URL typesetting
\usepackage{booktabs}       % professional-quality tables
\usepackage{amsfonts}       % blackboard math symbols
\usepackage{nicefrac}       % compact symbols for 1/2, etc.
\usepackage{microtype}      % microtypography
\usepackage{xcolor}         % colors
\usepackage[dvipsnames]{xcolor} % more color names
\usepackage{graphicx}
\usepackage{animate}        % for 360 video in teaser
\usepackage{subcaption}
\usepackage{tabularx}
\usepackage{makecell}
\usepackage{adjustbox}
\usepackage{setspace}
\newcolumntype{M}[1]{>{\centering\arraybackslash}m{#1}}
\usepackage{float}
\usepackage{tikz}
\usetikzlibrary{positioning,shapes,arrows}
\usepackage{amsmath,amsfonts,bm, bbm,leftindex}
\usepackage{multirow}
\usepackage{comment}
\usepackage{gensymb}
\usepackage{lipsum}
\usetikzlibrary{arrows.meta, positioning, fit}
\usepackage[para]{threeparttable}
\usepackage{tikz}
\usetikzlibrary{tikzmark}

\def\eqref#1{equation~\ref{#1}}
\def\1{\bm{1}}

\DeclareMathAlphabet{\mathsfit}{\encodingdefault}{\sfdefault}{m}{sl}
\SetMathAlphabet{\mathsfit}{bold}{\encodingdefault}{\sfdefault}{bx}{n}
\makeatletter
\let\save@mathaccent\mathaccent
\newcommand*\if@single[3]{%
  \setbox0\hbox{${\mathaccent"0362{#1}}^H$}%
  \setbox2\hbox{${\mathaccent"0362{\kern0pt#1}}^H$}%
  \ifdim\ht0=\ht2 #3\else #2\fi
  }
\newcommand*\rel@kern[1]{\kern#1\dimexpr\macc@kerna}
\newcommand*\widebar[1]{\@ifnextchar^{{\wide@bar{#1}{0}}}{\wide@bar{#1}{1}}}
\newcommand*\wide@bar[2]{\if@single{#1}{\wide@bar@{#1}{#2}{1}}{\wide@bar@{#1}{#2}{2}}}
\newcommand*\wide@bar@[3]{%
  \begingroup
  \def\mathaccent##1##2{%
    \let\mathaccent\save@mathaccent
    \if#32 \let\macc@nucleus\first@char \fi
    \setbox\z@\hbox{$\macc@style{\macc@nucleus}_{}$}%
    \setbox\tw@\hbox{$\macc@style{\macc@nucleus}{}_{}$}%
    \dimen@\wd\tw@
    \advance\dimen@-\wd\z@
    \divide\dimen@ 3
    \@tempdima\wd\tw@
    \advance\@tempdima-\scriptspace
    \divide\@tempdima 10
    \advance\dimen@-\@tempdima
    \ifdim\dimen@>\z@ \dimen@0pt\fi
    \rel@kern{0.6}\kern-\dimen@
    \if#31
      \overline{\rel@kern{-0.6}\kern\dimen@\macc@nucleus\rel@kern{0.4}\kern\dimen@}%
      \advance\dimen@0.4\dimexpr\macc@kerna
      \let\final@kern#2%
      \ifdim\dimen@<\z@ \let\final@kern1\fi
      \if\final@kern1 \kern-\dimen@\fi
    \else
      \overline{\rel@kern{-0.6}\kern\dimen@#1}%
    \fi
  }%
  \macc@depth\@ne
  \let\math@bgroup\@empty \let\math@egroup\macc@set@skewchar
  \mathsurround\z@ \frozen@everymath{\mathgroup\macc@group\relax}%
  \macc@set@skewchar\relax
  \let\mathaccentV\macc@nested@a
  \if#31
    \macc@nested@a\relax111{#1}%
  \else
    \def\gobble@till@marker##1\endmarker{}%
    \futurelet\first@char\gobble@till@marker#1\endmarker
    \ifcat\noexpand\first@char A\else
      \def\first@char{}%
    \fi
    \macc@nested@a\relax111{\first@char}%
  \fi
  \endgroup
}
\makeatother
\definecolor{darkred}{rgb}{0.7, 0.0, 0.0}
\definecolor{nvidia_green}{RGB}{118, 185, 0}

\usepackage{pifont}
\usepackage[nameinlink]{cleveref}
\crefname{equation}{Eq.}{Eqs.}
\crefname{figure}{Fig.}{Figs.}
\crefname{section}{Sec.}{Sec.}
\crefname{appendix}{App.}{App.}
\crefname{table}{Tab.}{Tabs.}
\crefname{algorithm}{Algo}{Algo}
\crefname{thm}{Thm}{Thm}
\Crefname{thm}{Thm}{Thm}
\crefname{prop}{Prop}{Prop}

\newcommand{\crefnames}[3]{%
  \@for\next:=#1\do{%
    \expandafter\crefname\expandafter{\next}{#2}{#3}%
  }%
}

\title{CuSfM: CUDA-Accelerated Structure-from-Motion}

\author{Jingrui Yu$^1$, Jun Liu$^1$, Kefei Ren$^1$, Joydeep Biswas$^1$, Rurui Ye$^1$, Keqiang Wu$^1$, Chirag Majithia$^1$, Di Zeng$^{1*}$\\
$^1$NVIDIA\\
$^*$Project Lead\\
\{jingruiy, junli, kefeir, jbiswas, ruruiye, keqiangw, cmajithia, dizeng\}@nvidia.com}

\begin{abstract}
Efficient and accurate camera pose estimation forms the foundational requirement for dense reconstruction in autonomous navigation, robotic perception, and virtual simulation systems. 
This paper addresses the challenge via cuSfM, a CUDA-accelerated offline Structure-from-Motion system that leverages GPU parallelization to efficiently employ computationally intensive yet highly accurate feature extractors, generating comprehensive and non-redundant data associations for precise camera pose estimation and globally consistent mapping. 
The system supports pose optimization, mapping, prior-map localization, and extrinsic refinement. It is designed for offline processing, where computational resources can be fully utilized to maximize accuracy. Experimental results demonstrate that cuSfM achieves significantly improved accuracy and processing speed compared to the widely used COLMAP method across various testing scenarios, while maintaining the high precision and global consistency essential for offline SfM applications. 
The system is released as an open-source Python wrapper implementation, PyCuSfM, available at \url{https://github.com/nvidia-isaac/pyCuSFM}, to facilitate research and applications in computer vision and robotics.
\end{abstract}

\begin{document}

\maketitle

\abscontent

\section{Introduction}
Structure-from-Motion (SfM), a well-established technique in computer vision, refers to the process of reconstructing accurate three-dimensional structures by analyzing the motion and viewpoints of a sequence of images. SfM finds extensive applications across diverse domains, including surveying and mapping, autonomous navigation, cultural heritage preservation, robotic perception, and virtual simulation systems. By enabling accurate 3D reconstruction from images, SfM supports diverse tasks from large-scale topographic mapping to fine-grained architectural documentation and robotic localization and perception.

\begin{figure}[!ht]
    \centering
    \includegraphics[width=0.8\linewidth]{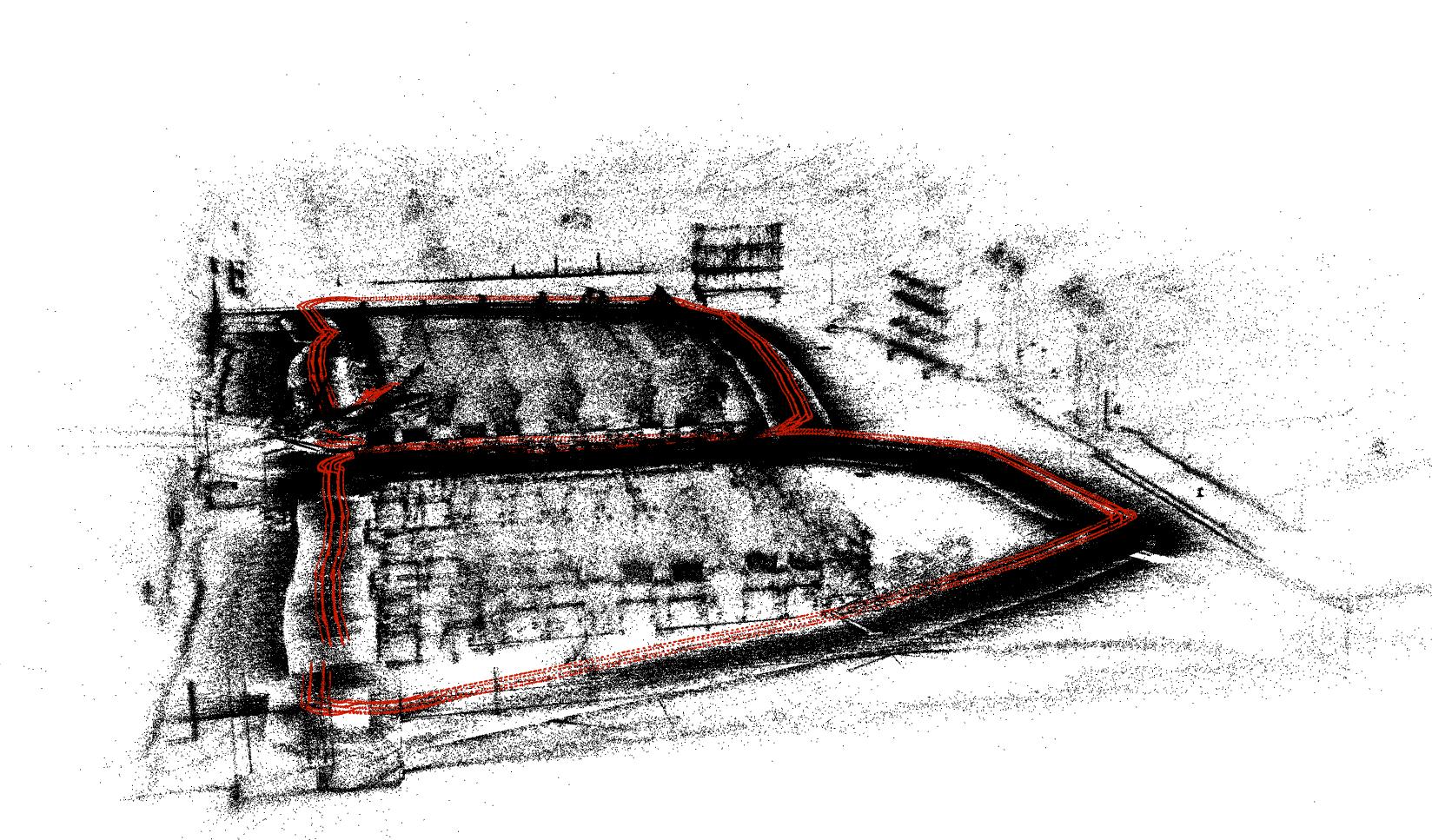}
    \caption{High-quality 3D reconstruction achieved by cuSfM: A dense point cloud of an indoor meeting room containing over 1.4 million 3D landmarks.}
    \label{fig:dense_point_cloud}
\end{figure}

COLMAP (\cite{colmap}) remains the most widely adopted SfM framework, employing an incremental strategy that prioritizes robustness over computational efficiency. Recent approaches such as GLOMAP (\cite{glomap}) use global optimization strategies and show measured improvements on evaluated benchmarks. These conventional methods maintain a modular pipeline comprising feature extraction, matching, and mapping stages. The mapping phase involves solving large-scale nonlinear optimization problems with nonconvex constraints, inherently generating substantial computational overhead and prolonged processing durations. Furthermore, extracting high-quality low-dimensional structural representations from high-dimensional visual inputs poses significant challenges for traditional techniques.

Advances in deep neural networks have enabled end-to-end, data-driven Structure-from-Motion (SfM) approaches, such as ACE-Zero (\cite{acezero}) and Light3R-SfM (\cite{light3r}), which bypass traditional explicit feature extraction and data association by directly predicting camera poses and scene structures. Although these methods show promise in controlled environments, recent study (\cite{MASt3R}) demonstrates that ACE-Zero and similar approaches tend to perform well only in constrained settings, struggling with unordered image collections exhibiting large viewpoint and illumination variations. This limitation arises because the regressor networks rely on incremental propagation of information between closely related images, which hinders generalization to more diverse, unstructured real-world scenes. Furthermore, their scalability is often limited by memory consumption, restricting their application to relatively small spatial scales.

In both outdoor autonomous driving and indoor robotic mapping, datasets often exhibit complex spatial and temporal inconsistencies. Common challenges include globally aligning disjoint trajectories acquired by multiple platforms, registering newly captured sensor observations to preexisting maps, and refining inter-camera extrinsic parameters within multi-sensor ego-vehicle configurations. Addressing these issues is critical for ensuring globally consistent localization and accurate large-scale reconstruction.

To overcome these limitations, this paper presents \textcolor{nvidia_green}{\textbf{cuSfM}}, a GPU-accelerated Structure-from-Motion framework designed for enhanced versatility and precision. CuSfM leverages global optimization strategies to achieve computational efficiency while integrating iterative triangulation with bundle adjustment to maintain outlier robustness. The framework combines data-driven feature extractors with advanced matching algorithms, preserving a classical nonlinear optimization backbone for pose estimation and environment reconstruction. This hybrid architecture supports multiple operational modes including localization and extrinsic parameter calibration.

All computational components in cuSfM leverage GPU-accelerated parallel processing, achieving significant efficiency gains without compromising accuracy. Experimental evaluations on benchmark datasets demonstrate an order-of-magnitude runtime improvement compared to COLMAP. The framework is publicly available\footnote{\url{https://github.com/nvidia-isaac/pyCuSFM}}, providing researchers with advanced tools for computer vision and robotics applications.

As demonstrated in Fig.~\ref{fig:dense_point_cloud}, cuSfM achieves high-quality 3D reconstruction results in real-world indoor environments, generating dense point clouds with over 1.4 million 3D landmarks from a meeting room scene. This capability showcases the system's practical effectiveness in creating detailed environmental representations suitable for downstream applications such as neural network-based reconstruction and virtual simulation systems.

\section{System Architecture of cuSfM}
\begin{figure}[!ht]
    \centering
    \includegraphics[width=1\linewidth]{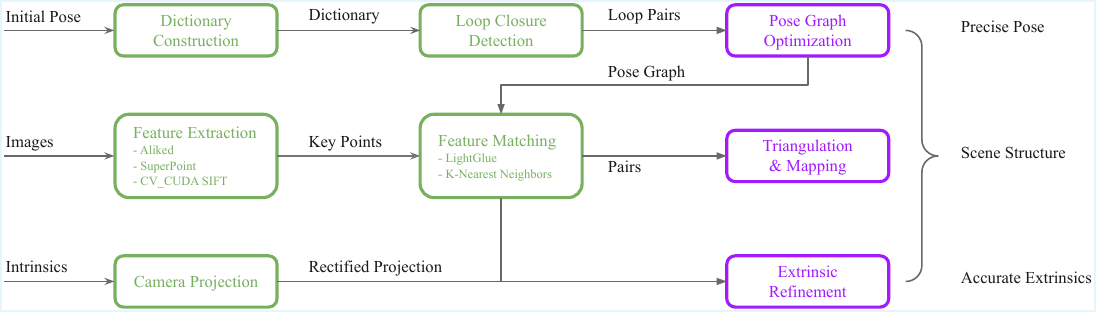}
    \caption{System architecture of cuSfM. The pipeline flows from left to right, comprising system inputs, non-redundant data association modules (highlighted in green), pose refinement modules based on data associations (highlighted in purple), and system outputs. The framework supports multiple operational modes including pose optimization and mapping, prior map-based localization, and extrinsic parameter refinement.}
    \label{fig:system_architecture}
\end{figure}

CuSfM adopts a modular architecture that enables flexible deployment across diverse mapping and localization scenarios, as illustrated in Fig.~\ref{fig:system_architecture}. The system accepts initial pose estimates, image sequences, and camera parameters as inputs, then processes them through specialized modules to produce precise pose, scene structure, and accurate extrinsics.

CuSfM operates as a vision-based pose optimizer that leverages prior trajectories. The system requires initial poses, time-continuous image sequences, and corresponding camera model information, including camera types, intrinsic parameters, and inter-camera extrinsic relationships, as inputs. The system outputs refined trajectories with improved accuracy, environmental structure information represented as sparse point clouds, and optimized inter-camera extrinsic parameters.

The first row of Figure~\ref{fig:system_architecture} demonstrates modules for global pose graph optimization. The system architecture incorporates three key components: dictionary construction as described in Section~\ref{cuvgl}, loop closure detection detailed in Section~\ref{loop_detection}, and pose graph optimization utilizing both loop closure and sequential pairs as presented in Section~\ref{pose_graph_optimization}. The framework also includes stereo frame pair pose estimation for loop closure frames, as detailed in Section~\ref{stereo_pose}. These modules facilitate the identification of frame adjacency relationships, enabling non-redundant feature matching. When loop closure frames are present and sequential frame relationships maintain accuracy, this approach yields fast and precise results. The framework further extends to multi-trajectory alignment applications.

The second row presents the classical SfM pipeline, comprising feature extraction, feature matching, triangulation, and Bundle Adjustment (BA) optimization. 
For feature extraction, the current implementation supports ALIKED (\cite{ALIKED2023}), CV\_CUDA SIFT and SuperPoint (\cite{superpoint}), with ALIKED demonstrating superior overall performance in experimental evaluations. 
For feature matching, the framework supports LightGlue (\cite{lightglue2023}) and classical k-NN matchers, with LightGlue showing superior performance. 
Both feature extraction and feature matching modules are designed for easy extension to accommodate future feature types, as detailed in Section~\ref{feature_extraction}. 
To mitigate computational redundancy and reduce noise interference from redundant data associations, cuSfM implements a pose graph-based view-graph construction strategy, which is elaborated in Section~\ref{view_graph_construction}.

The camera projection module in the third row serves as a fundamental component, supporting custom extensions for arbitrary camera types and their corresponding undistortion models, as discussed in Section~\ref{camera_projection}. The extrinsic optimization module extends the triangulation and BA modules, addressing the practical challenge of inter-camera extrinsic parameter perturbations, as detailed in Section~\ref{extrinsic}. Enabling extrinsic optimization often yields more accurate extrinsic parameters in such scenarios.

These modules collectively constitute cuSfM, with their interconnected nature enabling various operational modes including localization, crowdsourced mapping, and map updates. The following sections elaborate on the implementation details of each module.

Throughout the subsequent discussion, the following notation conventions are adopted: poses are denoted by $T$, while their homogeneous matrix representations are denoted by $\mathbf{T}$. The superscript notation $T_i^j$ represents the pose of frame $i$ expressed in coordinate system $j$, where $w$ denotes the world coordinate frame.

\section{Non-Redundant Data Association}
\label{non_redundant}

\subsection{GPU-Accelerated Robust Feature Extraction and Matching}
\label{feature_extraction}

Visual feature extraction and matching are essential to vision-based mapping, localization, and reconstruction tasks. Reliable keypoint detection and descriptor matching enable consistent pose estimation and accurate multi-view alignment, even in challenging environments with illumination changes, viewpoint variation, and temporal scene evolution. While classical algorithms such as BRIEF (\cite{brief}), ORB (\cite{orb2}), and SIFT (\cite{sift}) provided the groundwork for robust feature computation, recent learning-based methods—including SuperPoint and ALIKED—have demonstrated substantially improved robustness and adaptability. However, these advances often come at the cost of significant computational demand, motivating the adoption of GPU acceleration for real-time and large-scale applications.

CuSfM incorporates a modular, GPU-accelerated feature processing pipeline designed for flexibility and future extensibility. Feature extraction, description, and matching operations are encapsulated as CUDA modules that follow a standardized interface, supporting both classical and deep learning–based algorithms. This modularity allows easy substitution, optimization, or augmentation of individual components without affecting the system as a whole. By managing the entire feature pipeline on device memory, cuSfM avoids host-device transfer bottlenecks and supports efficient, large-scale visual processing.

The system provides optimized GPU implementations of three distinct feature extraction methods. 

\textbf{ALIKED} serves as a learning-based detector and descriptor, offering precise sub-pixel localization and robustness against severe viewpoint and appearance changes. 

\textbf{SuperPoint}, the most popular learning-based feature extraction method, provides reliable keypoint detection and descriptor computation through self-supervised training. 

\textbf{CV\_CUDA SIFT} provides GPU-accelerated extraction and description of classical features for interpretable and stable keypoint computation. 

For feature matching, the system employs 

\textbf{LightGlue}, an efficient deep matching method that implements adaptive pruning and co-visibility priors, scalable to dense correspondence settings. 

This modular architecture enables seamless integration of emerging feature extraction and matching approaches.

To evaluate and select optimal components, all feature extraction and matching algorithms were benchmarked on the two-view relative pose estimation task, using the protocol established by Roma (\cite{edstedt2024roma}). All experiments were conducted on a GeForce RTX 3080 GPU. The evaluation encompassed state-of-the-art methods including X-Feat (\cite{potje2024cvpr}), DISK (\cite{disk}), and MSDP (\cite{lipson2024multi}), measuring both processing time and pose estimation accuracy reported as Area Under the Curve (AUC) at angular error thresholds of 5°, 10°, and 20° (see Table~\ref{tab:two_view_benchmark}). Learning-based techniques such as Mast3r (\cite{mast3r_arxiv24}) and RoMa achieved the highest AUC(5°) scores (47\%), albeit with the longest runtimes (0.6–2.37 s per image pair). In comparison, ALIKED+LightGlue produced near-peak accuracy (45\% AUC at 5°) at a substantially lower cost (0.265 s per pair). Traditional SIFT+NN offered 38\% AUC at 5° with a 0.756 s per pair processing time.

Based on these comprehensive evaluation results, ALIKED is selected as the default feature extraction method and LightGlue as the default matching method within the cuSfM pipeline. Their combination delivers high accuracy and efficiency, forming a highly scalable foundation for large-scale SfM and visual SLAM, while the modular implementation remains adaptable to new feature extraction and matching developments.

\begin{table}[!ht]
    \centering
    \caption{Comparison of feature extraction and matching on two-view relative pose estimation}
    \label{tab:two_view_benchmark}
    \begin{tabular}{lcccc}
    \toprule
    \multirow{2}{*}{\textbf{Method}} & \multirow{2}{*}{\textbf{Time (s)}} & \multicolumn{3}{c}{\textbf{Accuracy (AUC)}} \\
    \cmidrule(lr){3-5}
    & & \textbf{5°} & \textbf{10°} & \textbf{20°} \\
    \midrule
    SIFT+NN & 0.756 & 38\% & 56\% & 70\% \\
    SuperPoint+LightGlue & 0.330 & 38\% & 59\% & 75\% \\
    X-Feat+LightGlue & 0.270 & 39\% & 59\% & 75\% \\   
    DISK+LightGlue & 0.390 & 41\% & 60\% & 75\% \\
    MSDP & 0.520 & 34\% & 51\% & 65\% \\
    Mast3r & 0.600 & \textcolor{nvidia_green}{\textbf{47\%}} & \textcolor{nvidia_green}{\textbf{64\%}} & \textcolor{nvidia_green}{\textbf{78\%}} \\
    RoMa & 2.370 & \textcolor{nvidia_green}{\textbf{47\%}} & \textcolor{nvidia_green}{\textbf{64\%}} & \textcolor{nvidia_green}{\textbf{78\%}} \\
    ALIKED+LightGlue & \textcolor{nvidia_green}{\textbf{0.265}} & 45\% & 63\% & 77\% \\
    \bottomrule
    \end{tabular}
\end{table}

To address the computational overhead of the original Python implementations (about 200 ms per image pair) of ALIKED and LightGlue, TensorRT-optimized engines based on \texttt{C++} were employed, reducing inference time to approximately 20 ms while maintaining matching quality.

The GPU-accelerated feature processing pipeline adopts a modular architecture that facilitates integration of emerging algorithms and supports multiple backends including CUDA-accelerated SIFT through CV\_CUDA SIFT and SuperPoint. This extensible framework enables flexible adaptation to diverse application scenarios while maintaining consistent computational efficiency.

\subsection{Bag of Words Dictionary Building}
\label{cuvgl}
The mapping system implements loop detection and relocalization through Bag-of-Words (BoW) techniques. Rather than utilizing generic shared dictionaries, the system constructs environment-specific vocabularies that capture unique visual characteristics of the input datasets, thereby enhancing discriminative power. The construction process aggregates local visual feature descriptors to form a hierarchical vocabulary tree, where internal nodes represent coarse visual pattern clusters and leaf nodes correspond to individual vocabulary words. This multi-level organization enables efficient visual information encoding and retrieval for robust loop closure detection.

The BoW vector computation requires a visual vocabulary specifically trained for the deployed descriptor type. While prior research suggests the efficacy of general-purpose dictionaries trained on diverse datasets, empirical evaluation demonstrates superior performance with environment-specific vocabularies that accurately model local visual word distributions. The vocabulary construction employs recursive K-means clustering to generate a hierarchical tree structure. Starting from the root level, the algorithm iteratively partitions descriptors into K clusters, with each subsequent level further subdividing the data until reaching the desired tree depth, where leaf nodes define the final vocabulary words. To expedite the vocabulary-building process, cuSfM employs a multi-threaded strategy, achieving speeds up to ten times faster than conventional single-threaded K-Means clustering.   

Traditional K-means based vocabulary construction methods require loading all feature descriptors into memory simultaneously, creating a bottleneck for large-scale applications. To address this limitation, cuSfM implements an incremental vocabulary construction approach based on the Balanced Iterative Reducing and Clustering using Hierarchies (BIRCH) algorithm (\cite{birch}). BIRCH processes data incrementally by maintaining a Clustering Feature (CF) tree that stores three essential cluster attributes: object count (N), linear sum (LS), and squared sum (SS). These statistical quantities enable efficient cluster updates and radius calculations through simple additive operations, facilitating dynamic clustering as new data arrives.

The integration of new image features into the vocabulary CF tree follows a root-to-leaf traversal pattern, selecting intermediate nodes that minimize feature distance. The system implements adaptive node splitting when leaf radii or internal node capacities exceed predefined thresholds. A global refinement step utilizing weighted hierarchical K-means clustering ensures clustering quality. The BIRCH-based vocabulary builder extends conventional approaches by maintaining comprehensive cluster statistical quantities, including centroid descriptors, sample counts, and second-order information.

The unified data structure supports both BIRCH and K-means based vocabulary construction, enabling vocabulary extension and updates beyond the initial construction phase. This adaptability proves particularly valuable for map coverage extension and dynamic environments where visual appearances evolve over time. 

\subsection{Loop Detection}
\label{loop_detection}

\begin{figure}[!ht]
    \centering
    \includegraphics[width=0.8\linewidth]{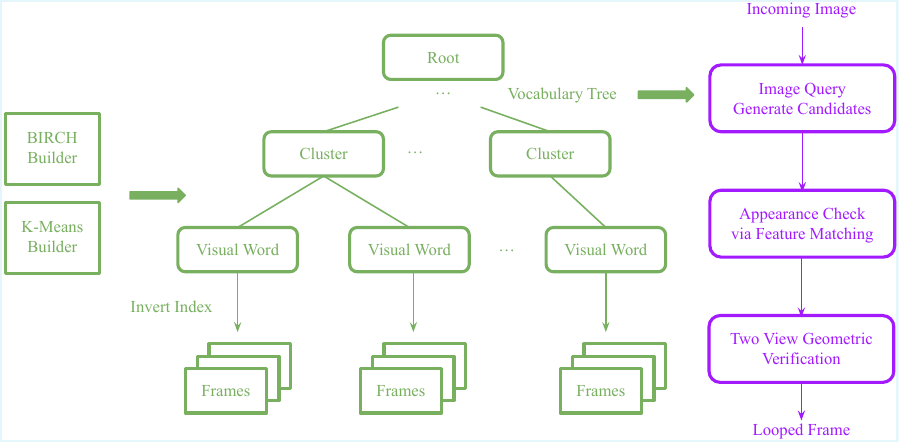}
    \caption{Illustration of the vocabulary tree construction and loop detection process. The left green panel shows the hierarchical structure of the vocabulary tree, where feature descriptors are clustered into visual words. The right purple panel demonstrates the loop detection pipeline.}
    \label{fig:voc}
\end{figure}

The complete image retrieval algorithm is a multi-stage process designed for efficient and accurate matching between incoming frames and stored keyframes. As shown in Fig.~\ref{fig:voc}, it consists of vocabulary tree construction and loop detection. First, a vocabulary tree is built from input image sequences to encode visual information into a hierarchical dictionary file. During mapping, each keyframe is added to a database augmented with an inverted index (loop index) that links visual words to keyframes. This structure enables fast querying across all keyframes, facilitating rapid identification of relevant frames during retrieval.

The second stage focuses on candidate map frame selection. For every new incoming image, the system converts the local feature descriptors extracted from the keyframe into a BoW vector. This vector acts as a global and compact representation of the keyframe's visual content. Utilizing the previously constructed lookup index, the algorithm computes a correlation score for each keyframe in the map by taking the inner product of the BoW vectors. This scoring mechanism effectively measures the similarity between the incoming image and each stored keyframe, enabling the system to identify and select the most relevant keyframes as candidates for global localization. The outcome of this stage is a ranked list of candidate keyframes, which are then forwarded to the next phase for more detailed geometric verification.

The third stage involves local feature matching between the selected candidate keyframes and the incoming frame. In this step, the system establishes correspondences by comparing the appearance of local image features, typically by calculating the L2 distance between the descriptors of potential feature pairs. To further enhance the robustness and accuracy of the matching process, several advanced techniques are employed. These include forward and backward matching, which ensures mutual consistency between matches, as well as the application of cost ratio thresholding, which compares the best and second-best matches to filter out ambiguous correspondences. Such measures are essential for increasing the success rate of feature matching, particularly in challenging or cluttered environments.

The final stage of the image retrieval algorithm is geometry validation, which serves as a critical filter to ensure the geometric consistency of the matched feature pairs. Each pair of matched features between the two images is subjected to a two-view fundamental matrix check, a process that determines whether the correspondence is consistent with the underlying epipolar geometry. Feature point pairs that satisfy this geometric constraint are classified as inliers, while those that do not are considered outliers and are discarded. The number of inliers resulting from this validation step serves as a key indicator of the reliability and success of loop frame detection, with a higher number of inliers generally signifying a more confident and accurate match.

It is important to note that the image retrieval algorithm described above is primarily responsible for identifying looped map frames that correspond to the incoming frame. However, the process of fully localizing the new frame within the map is completed by the relative pose estimation module, as detailed in Section \ref{stereo_pose}. This subsequent module refines the localization result by computing the precise spatial relationship between the matched frames, thereby enabling accurate and robust map-based localization within the visual mapping system.

\subsection{View-Graph Construction via Pose Graph Priors}
\label{view_graph_construction}
The view-graph construction module aims to establish reliable local feature associations between keyframes by selecting an optimal set of image pairs from the input data. These inter-keyframe connections collectively form a view graph, which serves as the fundamental structure for solving the SfM problem. Traditional incremental SfM systems, such as COLMAP, employ sophisticated next-best-view selection algorithms to incrementally determine image pairs that are likely to provide useful geometric constraints based on co-visibility and other heuristics. In contrast, global SfM approaches face a unique challenge: frame co-visibility information cannot be readily obtained in advance since the entire image set is processed collectively. However, the availability of initial pose information in typical scenarios provides an opportunity to guide the selection of image pairs for establishing data associations.

\begin{figure}[!ht]
    \centering
    \includegraphics[width=0.8\linewidth]{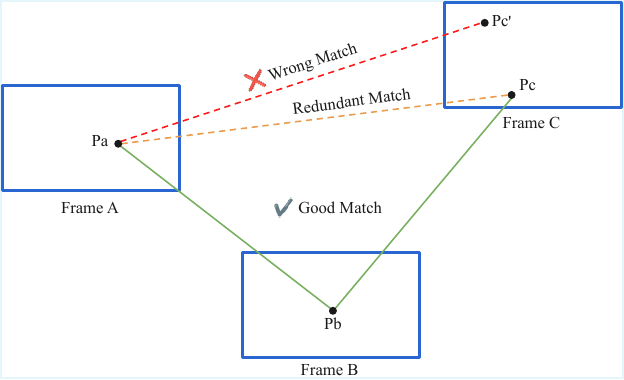}
    \caption{Illustration of the redundancy issue in feature matching. When Frame A matches with Frame B, and Frame B matches with Frame C, direct matching between Frames A and C becomes redundant as the correspondence can be established through Frame B. The optimal approach is to select only the minimal set of edges (A-B and B-C) that ensure robust triangulation while maintaining sufficient coverage.}
    \label{fig:vg}
\end{figure}

A conventional approach for generating candidate image matching pairs involves computing pose distances between keyframes and selecting pairs that exhibit proximity in both spatial position and orientation. The pose distance metric incorporates both translational and rotational components, measuring the spatial separation between camera positions and the angular difference in camera orientations, respectively. For each frame, spatially proximate candidates are efficiently identified using k-d tree-based nearest neighbor search, followed by filtering based on rotation distance to retain only viewpoint-compatible frames. While this approach requires predefined thresholds for translation and rotation distances as selection criteria, practical applications often necessitate threshold relaxation to include geometrically significant image pairs. This relaxation typically leads to a rapid increase in candidate pairs, causing the image matching step to become a computational bottleneck due to the inherent redundancy in the generated pairs.

As illustrated in Fig.~\ref{fig:vg}, the redundancy issue can be demonstrated through a three-frame scenario involving Frames A, B, and C. When Frame A matches with Frame B, and Frame B matches with Frame C, points in Frame A can be transitively associated with points in Frame C through Frame B. Direct matching between Frames A and C becomes redundant, particularly when the pose distance between A and B is smaller than that between A and C. Consider a point $P_a$ in Frame A matching with point $P_c$ in Frame C through an intermediate point $P_b$ in Frame B. Direct matching between Frames A and C either confirms the existing correspondence (providing no new information) or generates potentially conflicting matches. Therefore, including all three pairs—(A, B), (B, C), and (A, C)—in the feature matching process is inefficient. The optimal approach involves selecting only the minimal set of edges that ensure robust triangulation while maintaining sufficient coverage.

Based on the principle of maintaining complete yet non-redundant information sets (\cite{hierarchical}) and drawing inspiration from the Essential Graph concept in ORB-SLAM2 (\cite{orb2}), this work presents an alternative image pair selection method that leverages the inherent structure of the pose graph. The pose graph optimization framework represents keyframes as nodes and their spatial-temporal relationships as edges, comprising three distinct edge types: time-consecutive frame links, loop closure links, and extrinsic frame links. In video-based applications, consecutive frame links naturally exhibit high feature point co-visibility due to temporal and spatial proximity. Loop closure links, established through visual place recognition, connect revisited locations and are crucial for trajectory drift correction. Extrinsic frame links connect synchronized multi-camera captures, providing high-precision geometric constraints derived from calibration data. Experimental results demonstrate that constructing the view graph exclusively from pose graph edges achieves successful SfM reconstruction while significantly reducing computational overhead through non-redundant data association.

\subsection{Arbitrary Camera Type and Quantity Support}
\label{camera_projection}
While the pinhole camera model represents the most basic projection model, practical applications often require more sophisticated models to handle complex optical systems such as fisheye and panoramic cameras. In autonomous driving scenarios, additional distortion models become essential for cameras mounted behind windshields due to the optical effects of the glass medium. These applications necessitate various undistortion operations for images, feature points, and projection models.

The current implementation of cuSfM provides built-in support for two representative camera models: standard cameras with undistortion capabilities and F-theta cameras incorporating windshield models. The camera projection module adopts an extensible architecture that enables integration of additional camera models through the specification of bidirectional mapping functions between pixel coordinates and world coordinates. Furthermore, the camera projection module supports the conversion of distorted camera models to undistorted ones and the corresponding undistortion operations for images and feature points.

The camera projection modular design ensures that feature matching and triangulation-based mapping components operate independently of the underlying projection model, treating it as an abstract interface. This abstraction facilitates the incorporation of arbitrary camera models without modifications to the core reconstruction pipeline.

\section{Lightweight Pose Refinement}
\subsection{Stereo Relative Pose Estimation}
\label{stereo_pose}

In conventional approaches (\cite{Hartley:2003:MVG:861369}) to two-view geometry, the fundamental matrix check is commonly employed as a robust mechanism for eliminating outlier correspondences during the descriptor matching process between two image views. This geometric verification step ensures that only those feature matches which are consistent with the underlying epipolar geometry are retained for further analysis. By decomposing the fundamental matrix obtained from this procedure, it is possible to extract the relative rotation between the two camera views. However, a significant limitation of this method is that the relative translation between the two views can only be determined up to an unknown scale factor, meaning that the absolute distance between the cameras remains ambiguous.

To recover the missing scale information, previous state-of-the-art methods have typically relied on a combination of landmark triangulation and the Perspective-n-Point (SolvePnP) algorithm. In this framework, 3D landmarks are first triangulated from multiple views, and their scaled 3D coordinates are projected back into the image plane to estimate the full six degrees of freedom (6-DOF) pose of the image frames. While this approach is theoretically sound, it is prone to introducing errors in practice, primarily because accurate triangulation of 3D landmark coordinates is a non-trivial task, especially in the presence of noise, limited baseline, or poor feature localization. Moreover, it has been observed that 3D landmarks which are beneficial for accurate translation estimation may not necessarily be optimal for rotation estimation, and vice versa. This has led to the adoption of various heuristic strategies for selecting subsets of 3D landmarks tailored to either translation or rotation estimation. Unfortunately, these heuristics often lack solid theoretical justification and can be cumbersome to implement effectively in real-world systems.

This work presents a novel algorithm that directly estimates the translation scale using only 2D observations, thereby circumventing the need for both landmark triangulation and the SolvePnP procedure. The proposed approach estimates the full 6-DOF relative pose between a given map frame and the left image of an incoming stereo frame, relying solely on image-based information. The algorithm consists of three principal steps, as illustrated in the accompanying diagram. The first step performs a two-view geometric check and recovers the initial pose from the essential matrix, which is derived from the geometric relationships established during feature matching. The second step estimates the translation scale by leveraging the pose relationships among three views, effectively resolving the scale ambiguity that plagues traditional two-view methods. Finally, the third step refines the estimated pose through a joint optimization process, which integrates all available geometric constraints to produce a highly accurate and robust 6-DOF pose estimate. By avoiding the pitfalls associated with 3D landmark triangulation and heuristic landmark selection, this algorithm provides a theoretically grounded and practically efficient solution for relative pose estimation in visual mapping and localization systems.

\paragraph{Two-view geometry check and pose recovery} The initial step in the proposed methodology involves performing a comprehensive two-view geometric verification and estimating the rotation and translation direction for three distinct pairs of frames: the left frame and a selected map frame, the right frame and the same map frame, and the left and right frames themselves. In scenarios where the relative pose between the left and right frames is known with high accuracy from camera calibration, the two-view geometric check between the left and right frames can be omitted to streamline the process. For the remaining pairs, the two-view geometric check is carried out by identifying inlier 2D feature correspondences that are consistent with the underlying epipolar geometry. From these inlier matches, the essential matrices corresponding to the left/map frame pair and the map/right frame pair are computed. By decomposing each essential matrix, it is possible to estimate the rotation component of the relative pose between the respective pairs of frames. However, while the direction of translation can also be inferred from this decomposition, the magnitude of the translation remains indeterminate due to the inherent scale ambiguity associated with the essential matrix. Thus, at this stage, only the direction of translation can be reliably recovered rather than its absolute value.

\paragraph{Initial estimation of the translation scale}

The relative pose from the map frame to the left frame and from the right frame to the map frame are represented by $T_{left}^{map}$ and $T_{map}^{right}$, then the right frame to the left frame relative pose can be calculated through Eq.~\ref{eq:linear_fit}.

\begin{equation}
    \label{eq:linear_fit}
    T_{left}^{right} = T_{left}^{map} \cdot T_{map}^{right},
\end{equation}

where $T_{left}^{map}$ contains rotation, translation direction and a translation scale factor for the transformation from right frame to left frame. $T_{map}^{right}$ contains similar information from right to the map frame. $T_{left}^{right}$ is from calibration, the translation scale is calibrated. From the Eq.~\ref{eq:linear_fit}, the two translation scale factors can be estimated using simple linear fitting methods or other methods. 

\paragraph{Pose refine through joint optimization}

Once the scale factor for the translation component of the relative pose has been estimated, it becomes possible to formulate an initial guess for the six degrees of freedom (6-DOF) transformation from the left frame to the map frame. However, this preliminary estimate can be further improved through a dedicated optimization procedure. Drawing inspiration from the approach described in SOFT2 (\cite{soft2}), this optimization can be viewed as a generalization of the conventional two-view geometric framework, extending it to incorporate information from multiple image pairs. The core objective of the optimization is to minimize the aggregate Sampson distance, which quantifies the discrepancy between a point and its corresponding epipolar line, across all matched points for all relevant image pairs.

Suppose the signed distance between an image point in homogeneous coordinates x and a line $l=[l_1,l_2,l_3]^T$ as
\begin{equation}
    \label{eq:dist-point_to_line}
    d(x,l) = \frac{l^Tx}{\sqrt{l_1^2+l_2^2}}.
\end{equation}
We can define the Sampson distances for two images with matched feature points $(x_i, x_i')$ for $i=1,..,N$, $N$ is totally number of matched features, as
\begin{equation}
    \label{eq:image_pair_cost}
    S = \sum_{i} [d^2(x_i,l_i')+d^2(x_i',l_i)],
\end{equation}
where $E$ is essential matrix, $l_i'=E^Tx_i'$ and $l_i=Ex_i$ are epipolar lines associated to point $x_i'$ and $x_i$. Mathematically, the optimization seeks to minimize the sum of the sampson distance $S(L,M)$ of the left and map frame pair and $S(M,R)$ for the map and right frame pair:
      \begin{equation}
    \label{eq:three_image_cost}
    \min [S(L,M)+S(M,R)].
    \end{equation}

By jointly considering these distances, the optimization leverages geometric constraints from multiple perspectives, thereby enhancing the accuracy and robustness of the estimated pose.

It is essential to emphasize that the success of this optimization critically depends on the quality of the feature correspondences used. To maximize estimation accuracy, all outlier matches must be rigorously eliminated during the initial geometric verification step. By ensuring that only inlier correspondences contribute to the optimization, the procedure effectively refines the 6-DOF pose estimate, resulting in a more precise and reliable alignment between the left frame and the map frame.

\subsection{Pose Graph Optimization}
\label{pose_graph_optimization}
A pose graph consists of pose nodes and edges, where pose nodes represent camera poses and edges represent constraint relationships between camera poses, as illustrated in the Fig.~\ref{fig:pg}. The edges can be categorized into three types:

\begin{figure}[!ht]
    \centering
    \includegraphics[width=0.8\linewidth]{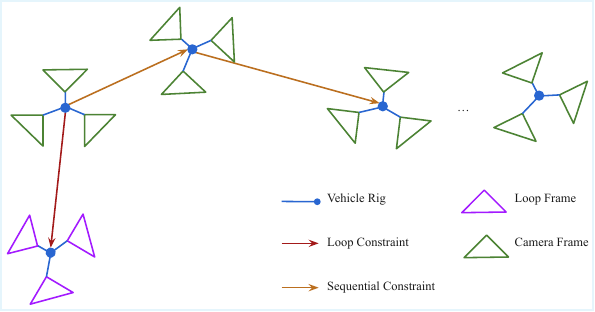}
    \caption{Illustration of the pose graph structure. Camera poses are represented as triangular nodes, with sequential constraints shown as yellow arrows and loop closure constraints as red arrows. Loop closure frames are highlighted in purple triangles, while blue lines connect three camera frames to form a vehicle rig.}
    \label{fig:pg}
\end{figure}

\begin{enumerate}
    \item Sequential constraints, derived from the initial trajectory. These constraints can be configured to establish relationships with several preceding and succeeding frames, expressed as,
    \begin{equation}
    \label{eq:seq}
    e_c^{(i,k)} =\|\mathbf{T}_i^w \cdot (\mathbf{T}_{i+k}^{w})^{-1} - \mathbf{T}_i^{i+k}\|^2.
    \end{equation}
    Empirical evidence suggests that constructing constraints only with adjacent frames ($k = 1$) is typically sufficient. Eq.~\ref{eq:seq} serves as the foundation for subsequent relative pose constraints in our framework.

    \item Loop closure constraints, derived from loop detection results. Upon detecting loop closure frames, the stereo relative pose estimation module computes the relative pose between loop closure frames and establishes corresponding constraints, expressed as $e_l = \operatorname{SRE}(F_i, F_j)$, where $F_j$ is the loop closure frame of $F_i$, and $\operatorname{SRE}$ represents the stereo relative pose estimation module (refer to Section~\ref{stereo_pose}).
    \item Extrinsic constraints, derived from the relative pose relationships between cameras within the same vehicle rig, which establish inter-camera constraints,
   \begin{equation}
    \label{eq:extrinsic}
    e_e^{(i,r)} =\|\mathbf{T}_i^w \cdot (\mathbf{T}_{r}^{w})^{-1} - \mathbf{T}_i^{r}\|^2,
    \end{equation}
    where $r$ represents another camera frame within the same vehicle rig as camera $i$.
\end{enumerate}

Using these three types of constraints, we construct a pose graph and optimize it using graph optimization algorithms. The objective function for pose graph optimization is formulated as:

\begin{equation}
    \min_{T_i} \sum_{i=1}^N e_i^T \Omega_i e_i,
\end{equation}

where $T_i$ represents the $i$-th pose node, $e_i$ denotes the corresponding constraint for the $i$-th pose node, and $\Omega_i$ is the weight matrix associated with the constraint.
The optimization objective aims to minimize both the constraint errors between pose nodes and their relative pose errors.

Two distinct configurations are proposed for pose graph nodes: \textit{camera frame} and \textit{vehicle rig} representations. In the \textit{camera frame} configuration, each individual camera is represented as a separate pose node, allowing for fine-grained optimization of individual camera poses. Alternatively, in the \textit{vehicle rig} configuration, the entire vehicle is treated as a single pose node, effectively incorporating multiple \textit{camera frames} into a unified representation. This latter approach inherently abstracts the extrinsic constraints between cameras, resulting in a more compact optimization problem with fewer variables. Experimental results demonstrate that the vehicle rig configuration often achieves superior pose estimation accuracy due to its reduced parameter space and implicit handling of inter-camera constraints.

\subsection{Iterative Triangulation and Mapping}
\label{iterative_triangulation}
Despite employing robust feature extractors and matchers to minimize outlier interference during the information representation stage, 
real-world systems inevitably encounter various intrinsic and extrinsic disturbances that prevent perfect information compression. 
To address this challenge, cuSfM employs an iterative triangulation and BA framework that alternates between 3D reconstruction and pose optimization to balance computational efficiency and reconstruction accuracy.

During the triangulation phase, multi-view correspondence tracks are constructed by organizing inter-frame feature matches into tracks, 
where each track represents observations of a single 3D point across multiple frames, $\text{track}_i = \{p_i^{(1)}, p_i^{(2)}, \ldots, p_i^{(k)}\}$, 
where $p_i^{(j)}$ denotes the projection of the $i$-th 3D point in the $j$-th frame. The track construction employs depth-first search over the feature matching graph to establish temporal correspondences.

For each constructed track, RANSAC-based triangulation is applied to estimate the 3D point coordinates. 
Successfully triangulated points and their corresponding projections are incorporated into the BA optimization, 
while remaining points in the track undergo iterative triangulation attempts until convergence or failure.

The Direct Linear Transform triangulation method shown as Eq.~\ref{eq:dlt} can operate in both image and normalized camera coordinates, supporting arbitrary nonlinear complex projection imaging processes,
\begin{equation}
\label{eq:dlt}
\mathbf{X} = \arg\min_{\mathbf{X}} \sum_{j=1}^{N} \|\mathbf{P}_j \mathbf{X} - \lambda_j \mathbf{p}_j\|^2,
\end{equation}
where $\mathbf{P}_j$ represents the camera projection matrix for frame $j$, and $\lambda_j$ denotes the homogeneous scaling factor. Furthermore, cuSfM also implements a midpoint-based triangulation approach (\cite{midTriangulation}). These diverse triangulation methods extend the applicability of cuSfM across various scenarios, ranging from challenging yet common small baseline angle scenarios where camera orientation aligns with motion direction, such as autonomous driving, to platforms utilizing multi-camera systems, such as robotics.

Successfully triangulated 3D points are incorporated into BA optimization through reprojection error constraints. For standard global shutter cameras, the reprojection error is formulated as:
\begin{equation}
e_r^{(i,j)} = \|\pi(\mathbf{K}_j \mathbf{T}_j^w \mathbf{X}_i) - \mathbf{p}_j^{obs}\|^2,
\end{equation}
where $\pi(\cdot)$ denotes the perspective projection function, $\mathbf{K}_i$ is the intrinsic matrix, $\mathbf{T}_i$ is the camera pose, and $\mathbf{p}_i^{obs}$ represents the observed feature point.

To accommodate real-world imaging conditions, the framework extends to rolling shutter cameras, where exposure-induced motion blur necessitates pose interpolation:
\begin{equation}
e_{rs}^{(i,j)} = \|\pi(\mathbf{K}_j \mathbf{T}_j^w(t_{scan}) \mathbf{X}_i) - \mathbf{p}_j^{obs}\|^2,
\end{equation}
where $\mathbf{T}_i(t_{scan})$ represents the interpolated pose at the scanline timestamp $t_{scan}$.

The keyframes associated with these projection points are also incorporated into BA optimization, with camera pose constraints constructed following the sequential frame constraint approach referenced in Eq.~\ref{eq:seq}. 
The distinguishing aspect lies in the BA stage, where arbitrary subsets of camera poses can be designated as fixed values. This functionality is achieved through the incorporation of absolute pose constraints:
\begin{equation}
\label{eq:abs}
e_a^{(i)} = \|T_i^w - T_{i,init}^{w}\|^2,
\end{equation}
where $T_{i,init}^{w}$ is the initial camera pose. This capability enables aligning new data to existing maps, facilitating either incremental updates with fixed existing maps or joint optimization incorporating existing map data. Therefore, cuSfM has the basis for implementing crowdsourced mapping.

Triangulation and BA optimization proceed in an alternating fashion. Upon completion of each BA optimization round, reprojection errors are computed based on updated 3D points and camera positions, with projection points exceeding the outlier threshold being removed, followed by continued triangulation attempts on remaining points in the tracks.

The iterative framework employs a two-stage optimization strategy. The first stage adopts relaxed outlier thresholds and robust loss functions to ensure rapid convergence.
The optimization objective combines reprojection errors with temporal consistency constraints,
\begin{equation}
\mathcal{L} = \sum_{i,j} \rho(e_r^{(i,j)}) + \lambda_c \sum_{k} e_c^{(i,k)} + \lambda_a \sum_{i} e_a^{(i)},
\end{equation}
where $\rho(\cdot)$ denotes a robust loss function and $e_c^{(k)}$ represents consecutive frame constraints.
The second stage applies stringent outlier criteria without loss functions to achieve high-precision results.

\subsection{Extrinsic Refinement}
\label{extrinsic}
Beyond camera pose optimization and environmental structure reconstruction, cuSfM incorporates a sophisticated extrinsic parameter refinement capability. This functionality is implemented through the vehicle rig configuration, where multiple cameras are treated as a unified rigid body, enabling simultaneous optimization of vehicle poses and camera-to-vehicle transformations.

In this formulation, each camera's pose is decomposed into two components: the vehicle pose in the world coordinate frame $T_v^w$ and the camera-to-vehicle extrinsic transformation $T_c^v$. This rigid-body formulation reflects physical system constraints by linking all cameras to a common mobile base through the composition, $T_c = T_c^v \cdot T_v^w$.

The factor graph for extrinsic refinement maintains the same structural elements as the basic optimization framework, incorporating reprojection error constraints, absolute pose constraints, and relative pose constraints to ensure both geometric accuracy and physical plausibility in multi-camera vehicle systems. The constituent error terms are modified to account for the vehicle rig configuration. Specifically, the reprojection error incorporates both camera extrinsics and vehicle pose:
\begin{equation}
e_{r}^{(i,j)} = \|\pi(\mathbf{K}_j \mathbf{T}_{c_j}^{v_j} \mathbf{T}_{v_j}^w \mathbf{X}_i) - \mathbf{p}_j^{obs}\|^2.
\end{equation}

Prior information about extrinsic parameters is incorporated into the factor graph through absolute pose constraints:
\begin{equation}
    e_{a}^{(i)} = \|T_i^v - T_{i,init}^{v}\|^2.
\end{equation}
This extension ensures that camera-to-vehicle transformations remain within reasonable bounds of their initial calibration values, which are typically obtained through manual measurement or prior calibration procedures. 
This approach prevents arbitrary drift in the vehicle coordinate system and preserves physical plausibility, while still permitting incremental refinement to correct for minor errors present in the initial calibration.

Additionally, the extrinsic constraints defined in Eq.~\ref{eq:extrinsic}, represented by $e_e^{(i,r)}$, preserve the known geometric relationships between cameras within the same rig—such as baselines and orientations—as established by the initial calibration. By enforcing rigidity across the camera group, it further reduces ambiguity and prevents degenerate solutions in which cameras could move independently in ways that are physically infeasible for the vehicle.

\section{Format Support}
CuSfM employs Protocol Buffers as the primary data serialization format for internal data definition and storage. The system utilizes JSON files for input and output of keyframe and camera information, while internal components such as feature point data, feature matching relationships, and pose graph constraints are stored in protobuf binary files. Additionally, module configuration parameters are defined using protobuf schemas and persisted as plain text files.

To facilitate interoperability with various open-source tools and evaluation frameworks, cuSfM supports multiple standard output formats:
\begin{itemize}
    \item \textbf{COLMAP\footnote{\url{https://colmap.github.io/format.html}}}: The system exports sparse point cloud maps using COLMAP's sparse reconstruction format, enabling visualization and further processing of the reconstructed 3D structure.
    \item \textbf{TUM\footnote{\url{https://cvg.cit.tum.de/data/datasets/rgbd-dataset/file_formats}}}: Optimized camera trajectories are exported in TUM format, facilitating quantitative evaluation using established tools such as evo and other trajectory analysis frameworks.
\end{itemize}

\section{Experiments}
This section presents comprehensive experiments to evaluate the proposed system from three main aspects: computational efficiency, accuracy performance, and multi-mode functionality demonstration.

\subsection{Experimental Setup}
All experiments were conducted on a workstation equipped with an Intel Core i9-9820X CPU @ 3.30GHz (10 cores, 20 threads) and an NVIDIA RTX 6000 Ada Generation GPU with 48GB VRAM. The system has Ubuntu 24.04 LTS as its operating system, with NVIDIA driver version 560.35.03 and CUDA 12.6 support. To ensure fair comparison, all evaluated methods were configured to utilize GPU acceleration capabilities on this identical hardware platform. The software environment includes PyCuSfM v0.1.8 as the proposed system, COLMAP v3.11.0 and GLMAP-GPU v1.1.0 for comparison. In addition, NVIDIA's DRIVE-SIM platform\footnote{\url{https://www.nvidia.com/en-us/use-cases/autonomous-vehicle-simulation/}} was leveraged to generate a small subset of simulated data with absolute ground truth for the experiments. This dataset, referred to as the SDG dataset, is designed to better represent the complexity of visual systems used in real-world production environments, which often consist of multiple camera systems of varying types. This dataset has been made publicly available\footnote{\url{https://github.com/nvidia-isaac/pyCuSFM/tree/main/data/SDG}} to facilitate further research and development in both academic and industrial settings. COLMAP and GLOMAP represent the most widely adopted SfM solutions in industrial applications, making them natural choices for comparative evaluation. While PyCuSfM primarily utilizes PyCuVSLAM (\cite{cuvslam}) for initial pose estimation, experiments were also conducted using ORB-SLAM2 (\cite{orb2}) as an alternative pose initializer to demonstrate the system's versatility.

\subsection{Computational Efficiency}
The computational efficiency evaluation was performed on the KITTI dataset (\cite{Geiger2012CVPR}). Table~\ref{tab:runtime_comparison} presents the average processing time per 100 frames for each pipeline stage across different methods. 

COLMAP and GLOMAP were configured to process monocular input data using their respective GPU-accelerated modes: COLMAP with sequential matcher mode and SIFT-GPU as the default feature detector and descriptor, and GLOMAP with its built-in GPU acceleration capabilities. CuSfM was evaluated with two different retrieval modes: the radius-based search mode, which utilizes all available data associations, and the view-graph mode, which employs a non-redundant data association strategy. Both modes utilize a TensorRT-accelerated implementation of ALIKED features. The radius-based search mode selects candidate frames within a 20-meter radius and 90-degree angular deviation from each frame's initial position. In contrast, the view-graph retrieval strategy leverages sequential information by selecting only the adjacent frames and loop closure frames as candidates. All evaluated systems utilize Ceres-Solver with SPARSE\_NORMAL\_CHOLESKY linear solver type for Bundle Adjustment optimization.

\begin{table}[!ht]
    \centering
    \caption{Average runtime comparison (seconds/100 frames) on KITTI sequences}
    \label{tab:runtime_comparison}
    \begin{tabular}{lcccc}
    \toprule
    \textbf{Pipeline Stage} & \textbf{COLMAP} & \textbf{GLOMAP} & \textbf{CuSfM} & \textbf{CuSfM} \\
    & & & \textbf{(radius)} & \textbf{(view-graph)} \\
    \midrule
    Feature Extraction    & 2.004  & \textcolor{nvidia_green}{\textbf{1.660}}  & 4.996 & 4.996 \\
    Feature Matching      & 3.690  & \textcolor{nvidia_green}{\textbf{3.262}}  & 76.600 & 37.370 \\
    Mapping              & 340.467 & 97.440 & 26.397 & \textcolor{nvidia_green}{\textbf{16.458}} \\
    \midrule
    \textbf{Total}       & 346.162 & 102.362 & 107.993 & \textcolor{nvidia_green}{\textbf{58.824}} \\
    \bottomrule
    \end{tabular}
\end{table}

Experimental results reveal that feature matching constitutes the most time-consuming stage in cuSfM. This computational overhead stems from two primary factors. First, the adoption of ALIKED features, while computationally more intensive than SIFT-GPU employed in COLMAP and GLOMAP, provides enhanced robustness in feature detection and matching. Second, the choice of retrieval strategy significantly impacts processing time, explaining the observed performance differences between the two matching modes. The radius-based strategy, which incorporates all geometrically constrained keyframes within a specified radius, necessitates approximately four times as many frame-to-frame matches compared to the view-graph approach. However, it is important to note that the view-graph strategy requires additional computational overhead for Bag of Words dictionary construction and search tree building, which the radius-based approach avoids. Consequently, the actual time difference between the two strategies is less pronounced than the disparity in the number of candidate frame pairs.

Despite the increased computational cost in these preliminary stages, cuSfM demonstrates remarkable efficiency in the mapping phase. Specifically, in view-graph mode, the mapping time is reduced by a factor of 20 (from 340.467s to 16.458s per 100 frames) compared to COLMAP, and by a factor of 6 compared to GLOMAP. This significant improvement leads to superior overall performance, with cuSfM's total processing time being only 16.931\% of COLMAP's (58.824s vs. 346.162s) and 57.467\% of GLOMAP's, even though GLOMAP already employs GPU acceleration and CUDA optimization. Such substantial performance gains have significant implications for industrial applications. Comparative analysis reveals that the radius search method identifies approximately seven times more image pair matches than the view-graph based approach, which inevitably leads to increased feature matching time.

The substantial improvement in mapping efficiency can be attributed to several architectural innovations. First, cuSfM implements a global reconstruction strategy, in contrast to COLMAP's incremental approach, which has proven suboptimal for industrial applications. Second, through the implementation of a non-redundant design principle detailed in Section \ref{non_redundant}, the system effectively utilizes coarse initial trajectory estimates, commonly available in practical deployment scenarios. This approach achieves optimal data association while minimizing redundant correspondence computations. The integration of these approaches, combined with iterative triangulation and GPU-accelerated bundle adjustment optimization, results in a highly efficient trajectory refinement and mapping pipeline.

\subsection{Accuracy Analysis}
Beyond computational efficiency, this section presents a comprehensive evaluation of pose accuracy through two comparative analyses: first, against state-of-the-art SfM methods, and second, relative to the input trajectory precision. Furthermore, to demonstrate the general applicability of trajectory refinement in the proposed system, experiments were conducted using initial trajectories from ORB-SLAM2, validating the system capability in optimizing poses from different sources.

The trajectory evaluation utilized the EVO toolkit (\cite{evo}) to compute the Absolute Trajectory Error (ATE). The evaluation metrics include Root Mean Square Error (RMSE), mean error, median error, standard deviation (STD), minimum error, and maximum error.

\subsubsection{Comparison with State-of-the-art SfM Methods}
\begin{figure}[!ht]
    \centering
    \includegraphics[width=\linewidth]{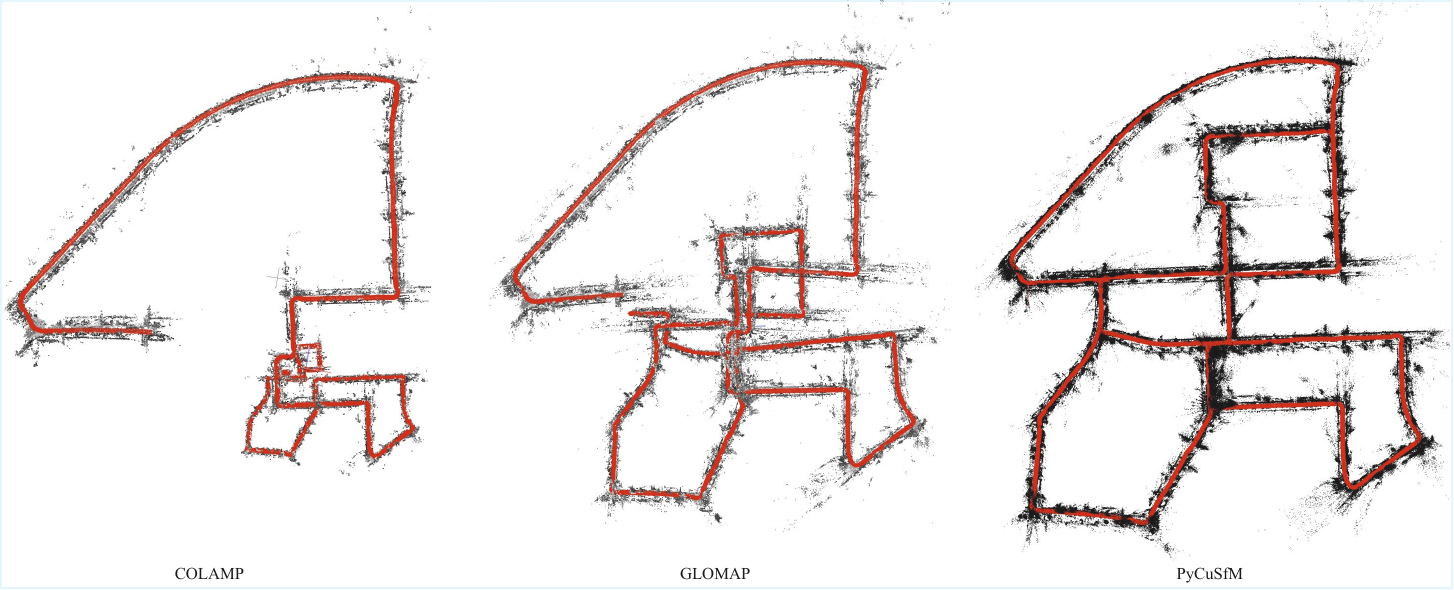}
    \caption{Trajectory comparison on KITTI 00 sequence.}
    \label{fig:kitti_00_reconstruction}
\end{figure}

Evaluation of both computational efficiency and trajectory accuracy was attempted by processing the KITTI dataset with COLMAP and GLOMAP. Due to the functional configuration limitations of these systems, which primarily support monocular input processing in their standard pipeline configurations, the evaluation was conducted using monocular image sequences extracted from the KITTI stereo dataset. However, as illustrated in Figure~\ref{fig:kitti_00_reconstruction}, which shows the reconstruction results of KITTI sequence 00, both systems exhibited significant limitations. Specifically, COLMAP and GLOMAP frequently failed to recover complete trajectories, and the reconstructed trajectories demonstrated inconsistent scale factors. Moreover, COLMAP was unable to reconstruct the entire trajectory within a single model, fragmenting the reconstruction into multiple disconnected components. 

These limitations may be attributed to two primary factors: the inability to leverage prior trajectory information and the potential unsuitability of their reconstruction strategies for autonomous driving scenarios. These observations highlight cuSfM's distinctive advantages in terms of both functionality and domain applicability, particularly in challenging automotive environments.

To provide a more comprehensive evaluation, additional experiments are conducted using the SDG dataset. The results are presented in Table~\ref{tab:sdg_accuracy}.

For these experiments, cuSfM was initialized with ground truth poses corrupted by Gaussian noise (maximum 3 meters in horizontal direction and 1 meter in vertical direction) to simulate realistic pose estimation errors encountered in practical applications. COLMAP and GLOMAP were evaluated using the front\_wide camera image sequences. The SDG dataset consists of relatively short sequences with simulated data, resulting in reduced systematic disturbances including image observation noise compared to real-world datasets. Consequently, COLMAP and GLOMAP achieved higher success rates on the SDG dataset compared to their performance on the KITTI dataset.

The experimental results demonstrate cuSfM's superior performance across most sequences. In Sequence 00, cuSfM achieved remarkable accuracy with RMSE of 0.040m, representing a 90.1\% improvement over COLMAP (0.406m) and 90.2\% improvement over GLOMAP (0.410m). Similarly, in Sequence 03, cuSfM significantly outperformed both baselines, achieving RMSE of 0.131m compared to COLMAP's 1.100m (88.1\% improvement) and GLOMAP's 0.675m (80.6\% improvement). Notably, COLMAP failed to complete trajectory reconstruction on Sequences 02 and 04, highlighting the limitations of traditional SfM approaches even in controlled simulation environments. In contrast, cuSfM successfully reconstructed all sequences with consistent accuracy, demonstrating its robustness and reliability.

Despite the favorable conditions of the simulated dataset, the results reveal significant performance gaps between cuSfM and traditional methods. The superior performance of cuSfM can be attributed to its ability to leverage prior trajectory information and its robust feature matching pipeline, which remains effective even when initial poses contain substantial noise.

\begin{table}[!ht]
    \centering
    \caption{Comparison of absolute trajectory error on SDG dataset sequences}
    \label{tab:sdg_accuracy}
    \begin{tabular}{lccccccc}
    \toprule
    \multirow{2}{*}{Sequence} & \multirow{2}{*}{Method} & \multicolumn{6}{c}{Absolute Trajectory Error (m)} \\
    \cmidrule(lr){3-8}
    & & RMSE & Mean & Median & STD & Min & Max \\
    \midrule
    \multirow{3}{*}{00} 
    & COLMAP & 0.406 & 0.372 & 0.395 & 0.164 & 0.043 & 0.721 \\
    & GLOMAP & 0.410 & 0.375 & 0.397 & 0.167 & 0.049 & 0.726 \\
    & CuSfM & \textcolor{nvidia_green}{\textbf{0.040}} & \textcolor{nvidia_green}{\textbf{0.038}} & \textcolor{nvidia_green}{\textbf{0.040}} & \textcolor{nvidia_green}{\textbf{0.013}} & \textcolor{nvidia_green}{\textbf{0.004}} & \textcolor{nvidia_green}{\textbf{0.059}} \\
    \midrule
    \multirow{3}{*}{01}
    & COLMAP & 0.053 & \textcolor{nvidia_green}{\textbf{0.040}} & \textcolor{nvidia_green}{\textbf{0.033}} & 0.034 & 0.003 & 0.156 \\
    & GLOMAP & 0.059 & 0.044 & 0.033 & 0.040 & \textcolor{nvidia_green}{\textbf{0.002}} & 0.185 \\
    & CuSfM & \textcolor{nvidia_green}{\textbf{0.046}} & 0.041 & 0.038 & \textcolor{nvidia_green}{\textbf{0.021}} & 0.014 & \textcolor{nvidia_green}{\textbf{0.091}} \\
    \midrule
    \multirow{3}{*}{02}
    & COLMAP & - & - & - & - & - & - \\
    & GLOMAP & 0.063 & 0.058 & 0.049 & 0.025 & 0.026 & 0.136 \\
    & CuSfM & \textcolor{nvidia_green}{\textbf{0.044}} & \textcolor{nvidia_green}{\textbf{0.038}} & \textcolor{nvidia_green}{\textbf{0.030}} & \textcolor{nvidia_green}{\textbf{0.021}} & \textcolor{nvidia_green}{\textbf{0.017}} & \textcolor{nvidia_green}{\textbf{0.107}} \\
    \midrule
    \multirow{3}{*}{03}
    & COLMAP & 1.100 & 1.013 & 1.075 & 0.428 & 0.307 & 1.924 \\
    & GLOMAP & 0.675 & 0.621 & 0.599 & 0.264 & 0.101 & 1.162 \\
    & CuSfM & \textcolor{nvidia_green}{\textbf{0.131}} & \textcolor{nvidia_green}{\textbf{0.094}} & \textcolor{nvidia_green}{\textbf{0.058}} & \textcolor{nvidia_green}{\textbf{0.091}} & \textcolor{nvidia_green}{\textbf{0.019}} & \textcolor{nvidia_green}{\textbf{0.431}} \\
    \midrule
    \multirow{3}{*}{04}
    & COLMAP & - & - & - & - & - & - \\
    & GLOMAP & 0.384 & 0.354 & 0.361 & 0.148 & 0.044 & 0.712 \\
    & CuSfM & \textcolor{nvidia_green}{\textbf{0.044}} & \textcolor{nvidia_green}{\textbf{0.040}} & \textcolor{nvidia_green}{\textbf{0.038}} & \textcolor{nvidia_green}{\textbf{0.020}} & \textcolor{nvidia_green}{\textbf{0.006}} & \textcolor{nvidia_green}{\textbf{0.098}} \\
    \midrule
    \multirow{3}{*}{05}
    & COLMAP & 0.839 & 0.765 & 0.806 & 0.345 & 0.106 & 1.321 \\
    & GLOMAP & \textcolor{nvidia_green}{\textbf{0.723}} & 0.656 & 0.692 & \textcolor{nvidia_green}{\textbf{0.304}} & 0.097 & \textcolor{nvidia_green}{\textbf{1.138}} \\
    & CuSfM & 0.907 & \textcolor{nvidia_green}{\textbf{0.518}} & \textcolor{nvidia_green}{\textbf{0.321}} & 0.745 & \textcolor{nvidia_green}{\textbf{0.075}} & 3.324 \\
    \bottomrule
    \end{tabular}
\end{table}

Based on our comprehensive evaluation, cuSfM demonstrates superior performance in dynamic scenarios, particularly in terms of trajectory completeness and environmental structure reconstruction. The robust performance and efficient processing make cuSfM particularly suitable for downstream tasks such as dense reconstruction using 3D Gaussian Splatting (3DGS) or Neural Radiance Fields (NeRF). These advantages position cuSfM as a promising solution for real-world applications requiring accurate and efficient 3D reconstruction.

\subsubsection{Trajectory Refinement Evaluation}
To evaluate the trajectory refinement performance, five challenging sequences from the KITTI dataset were selected for comparison with the original PyCuVSLAM trajectories. Additionally, to validate the advantages of non-redundant data association, two distinct strategies were employed and compared: a radius-based approach and a view-graph based approach for constructing data association candidate pairs, with results presented in Table~\ref{tab:trajectory_accuracy}.

The view-graph based data association method is detailed in Section~\ref{view_graph_construction}. In contrast, the radius-based strategy employs geometric constraints, selecting all frames within a specified distance from the current frame as data association candidates. For this experiment, the distance threshold was set to 20 meters with an angular deviation limit of 90 degrees.

\begin{table}[!ht]
    \centering
    \caption{Comparison of absolute trajectory error between cuSfM and CuVSLAM on KITTI sequences}
    \label{tab:trajectory_accuracy}
    \begin{tabular}{lccccccc}
    \toprule
    \multirow{2}{*}{Sequence} & \multirow{2}{*}{Method} & \multicolumn{6}{c}{Absolute Trajectory Error (m)} \\
    \cmidrule(lr){3-8}
    & & RMSE & Mean & Median & STD & Min & Max \\
    \midrule
    \multirow{3}{*}{00} 
    & CuVSLAM        & 2.022 & 1.798 & 1.780 & 0.925 & 0.236 & 4.806 \\
    & CuSfM-Radius & 1.712 & 1.467 & 1.261 & 0.881 & \textcolor{nvidia_green}{\textbf{0.148}} & 4.103 \\
    & CuSfM-View Graph  & \textcolor{nvidia_green}{\textbf{1.388}} & \textcolor{nvidia_green}{\textbf{1.189}} & \textcolor{nvidia_green}{\textbf{0.985}} & \textcolor{nvidia_green}{\textbf{0.716}} & 0.158 & \textcolor{nvidia_green}{\textbf{3.382}} \\
    \midrule
    \multirow{3}{*}{02}
    & CuVSLAM        & 3.127 & 2.906 & 2.723 & \textcolor{nvidia_green}{\textbf{1.155}} & 0.904 & 7.669 \\
    & CuSfM-Radius & 3.008 & 2.749 & 2.575 & 1.221 & 0.728 & \textcolor{nvidia_green}{\textbf{6.601}} \\
    & CuSfM-View Graph  & \textcolor{nvidia_green}{\textbf{2.899}} & \textcolor{nvidia_green}{\textbf{2.610}} & \textcolor{nvidia_green}{\textbf{2.422}} & 1.261 & \textcolor{nvidia_green}{\textbf{0.419}} & 6.714 \\
    \midrule
    \multirow{3}{*}{05}
    & CuVSLAM        & 2.028 & 1.808 & 1.437 & 0.919 & 0.786 & 5.081 \\
    & CuSfM-Radius & 1.623 & 1.433 & 1.348 & 0.762 & 0.285 & 3.529 \\
    & CuSfM-View Graph  & \textcolor{nvidia_green}{\textbf{1.174}} & \textcolor{nvidia_green}{\textbf{1.049}} & \textcolor{nvidia_green}{\textbf{1.034}} & \textcolor{nvidia_green}{\textbf{0.525}} & \textcolor{nvidia_green}{\textbf{0.277}} & \textcolor{nvidia_green}{\textbf{3.433}} \\
    \midrule
    \multirow{3}{*}{06}
    & CuVSLAM        & 1.202 & 1.055 & 0.907 & 0.577 & 0.234 & 2.855 \\
    & CuSfM-Radius & 0.813 & 0.769 & \textcolor{nvidia_green}{\textbf{0.665}} & 0.263 & 0.353 & 1.669 \\
    & CuSfM-View Graph  & \textcolor{nvidia_green}{\textbf{0.783}} & \textcolor{nvidia_green}{\textbf{0.741}} & 0.738 & \textcolor{nvidia_green}{\textbf{0.255}} & \textcolor{nvidia_green}{\textbf{0.112}} & \textcolor{nvidia_green}{\textbf{1.488}} \\
    \midrule
    \multirow{3}{*}{09}
    & CuVSLAM        & 3.322 & 2.505 & \textcolor{nvidia_green}{\textbf{1.243}} & 2.182 & 0.373 & 7.567 \\
    & CuSfM-Radius & 3.261 & 2.507 & 1.424 & 2.084 & \textcolor{nvidia_green}{\textbf{0.015}} & 7.977 \\
    & CuSfM-View Graph  & \textcolor{nvidia_green}{\textbf{2.040}} & \textcolor{nvidia_green}{\textbf{1.725}} & 1.268 & \textcolor{nvidia_green}{\textbf{1.089}} & 0.395 &  \textcolor{nvidia_green}{\textbf{3.882}}  \\
    \bottomrule
    \end{tabular}
\end{table}

The experimental results demonstrate that both the radius based strategy and the view-graph based strategy consistently achieve higher accuracy compared to the original PyCuVSLAM trajectories, validating the effectiveness of the proposed refinement approach. Specifically, the view-graph based cuSfM achieves significant improvements across all sequences, with RMSE reductions ranging from 7.3\% (Sequence 02) to 42.1\% (Sequence 05) compared to CuVSLAM. The most notable improvements are observed in Sequence 05, where the RMSE decreases from 2.028m to 1.174m, and in Sequence 09, where the maximum error is reduced by 48.7\% (from 7.567m to 3.882m). These substantial improvements validate the effectiveness of the proposed refinement approach. For practical applications where trajectory accuracy is critical, the integration of PyCuVSLAM and the proposed refinement system is recommended to achieve optimal results.

Comparative analysis of the two strategies reveals that the view-graph based cuSfM consistently demonstrates superior performance across all evaluation metrics. For instance, in Sequence 00, the view-graph approach reduces the RMSE by 18.9\% (from 1.712m to 1.388m) compared to the radius-based strategy, while also achieving better performance in mean error (1.189m vs 1.467m) and median error (0.985m vs 1.261m). Similar improvements are observed in other sequences, with the view-graph strategy showing particular advantages in reducing maximum errors, as evidenced by the 51.4\% reduction in Sequence 09 (from 7.977m to 3.882m). These experimental results suggest that in visual pose estimation problems, indiscriminately utilizing all available information is suboptimal, both in terms of computational efficiency and final accuracy. Instead, a more effective approach involves carefully selecting information to construct a complete and orthogonal information set for state estimation. While our experimental results cannot definitively prove that the view-graph based information set represents the ideal complete and orthogonal set - as the orthogonality between visual images is challenging to mathematically prove - our methodology provides a promising direction for future research in this domain.

\begin{table}[!ht]
    \centering
    \caption{Comparison of absolute trajectory error between cuSfM and ORB-SLAM2 on KITTI sequences}
    \label{tab:trajectory_comparison_orbslam}
    \begin{tabular}{lccccccc}
    \toprule
    \multirow{2}{*}{Sequence} & \multirow{2}{*}{Method} & \multicolumn{6}{c}{Absolute Trajectory Error (m)} \\
    \cmidrule(lr){3-8}
    & & RMSE & Mean & Median & STD & Min & Max \\
    \midrule
    \multirow{2}{*}{00} 
    & ORB-SLAM2 & 1.231 & 0.969 & \textcolor{nvidia_green}{\textbf{0.819}} & 0.760 & \textcolor{nvidia_green}{\textbf{0.018}} & 7.342 \\
    & CuSfM   & \textcolor{nvidia_green}{\textbf{0.979}} & \textcolor{nvidia_green}{\textbf{0.901}} & 0.839 & \textcolor{nvidia_green}{\textbf{0.384}} & 0.123 & \textcolor{nvidia_green}{\textbf{2.625}} \\
    \midrule
    \multirow{2}{*}{02}
    & ORB-SLAM2 & 5.480 & 4.557 & 3.645 & 3.043 & 0.824 & \textcolor{nvidia_green}{\textbf{12.497}} \\
    & CuSfM   & \textcolor{nvidia_green}{\textbf{5.444}} & \textcolor{nvidia_green}{\textbf{4.534}} & \textcolor{nvidia_green}{\textbf{3.575}} & \textcolor{nvidia_green}{\textbf{3.012}} & \textcolor{nvidia_green}{\textbf{0.810}} & 13.092 \\
    \midrule
    \multirow{2}{*}{05}
    & ORB-SLAM2 & 0.408 & 0.353 & 0.311 & 0.203 & 0.049 & \textcolor{nvidia_green}{\textbf{1.882}} \\
    & CuSfM   & \textcolor{nvidia_green}{\textbf{0.403}} & \textcolor{nvidia_green}{\textbf{0.351}} & \textcolor{nvidia_green}{\textbf{0.309}} & \textcolor{nvidia_green}{\textbf{0.198}} & \textcolor{nvidia_green}{\textbf{0.030}} & 1.882 \\
    \midrule
    \multirow{2}{*}{06}
    & ORB-SLAM2 & 0.638 & 0.599 & 0.551 & 0.223 & 0.094 & 1.032 \\
    & CuSfM   & \textcolor{nvidia_green}{\textbf{0.436}} & \textcolor{nvidia_green}{\textbf{0.436}} & \textcolor{nvidia_green}{\textbf{0.387}} & \textcolor{nvidia_green}{\textbf{0.175}} & \textcolor{nvidia_green}{\textbf{0.009}} & \textcolor{nvidia_green}{\textbf{0.788}} \\
    \midrule
    \multirow{2}{*}{09}
    & ORB-SLAM2 & 2.643 & 2.143 & 1.326 & 1.548 & 0.532 & 6.460 \\
    & CuSfM   & \textcolor{nvidia_green}{\textbf{1.523}} & \textcolor{nvidia_green}{\textbf{1.374}} & \textcolor{nvidia_green}{\textbf{1.158}} & \textcolor{nvidia_green}{\textbf{0.656}} & \textcolor{nvidia_green}{\textbf{0.458}} &  \textcolor{nvidia_green}{\textbf{3.010}} \\
    \bottomrule
    \end{tabular}
\end{table}

To demonstrate the general applicability of the refinement capabilities, additional experiments were conducted using ORB-SLAM2 as the trajectory initializer. The results, presented in Table~\ref{tab:trajectory_comparison_orbslam}, show consistent accuracy improvements across most sequences, further validating the system's refinement capabilities. Specifically, cuSfM achieves notable improvements in several key metrics:

In Sequence 00, while ORB-SLAM2 shows better performance in median error (0.819m vs 0.839m) and minimum error (0.018m vs 0.123m), cuSfM demonstrates significant advantages in overall trajectory consistency, reducing the RMSE by 20.5\% (from 1.231m to 0.979m) and the standard deviation by 49.5\% (from 0.760m to 0.384m). Most notably, the maximum error is reduced by 64.2\% (from 7.342m to 2.625m), indicating much better handling of challenging trajectory segments. Sequence 06 shows the most consistent improvements across all metrics, with cuSfM reducing the RMSE by 31.7\% (from 0.638m to 0.436m), mean error by 27.2\% (from 0.599m to 0.436m), and standard deviation by 21.5\% (from 0.223m to 0.175m). The maximum error is also reduced by 23.6\% (from 1.032m to 0.788m), demonstrating robust performance across the entire trajectory. In Sequence 09, cuSfM achieves substantial improvements in all error metrics, with a 42.4\% reduction in RMSE (from 2.643m to 1.523m), 35.9\% reduction in mean error (from 2.143m to 1.374m), and 57.6\% reduction in maximum error (from 6.460m to 3.010m). These improvements are particularly significant given the challenging nature of this sequence. While Sequences 02 and 05 show more modest improvements, with RMSE reductions of only 0.7\% and 1.2\% respectively, it's worth noting that these sequences already demonstrate relatively good performance from ORB-SLAM2. The consistent improvement across all metrics in these sequences, albeit small, still validates the robustness of the refinement approach.

Analysis of both experimental scenarios indicates that the refinement performance exhibits dependency on initial values, with the system achieving comparable error levels across different input trajectories. This behavior can be attributed to the non-convex nature of trajectory optimization and Structure-from-Motion problems, where initial values inherently influence the optimization outcome. The results demonstrate that cuSfM can effectively refine trajectories regardless of the initializer used, while maintaining or improving upon the accuracy of the input trajectories.

\subsection{Multi-Mode Functionality Demonstration}

\subsubsection{Extrinsic Refinement}

As introduced in Section \ref{extrinsic}, cuSfM extends beyond pose optimization to incorporate extrinsic parameter refinement capabilities. 
Our empirical findings demonstrate that decoupling extrinsic parameters from robot poses often leads to enhanced pose estimation accuracy. 
To evaluate the effectiveness of this approach, we conducted comprehensive experiments on the KITTI dataset, with results presented in Table \ref{tab:extrinsic_refinement}. For our baseline, we employed the view-graph based cuSfM configuration (denoted as cuSFM w.o. ER in the table), while the extrinsic refinement variant (cuSFM w/ ER) differs only in the final mapping step.

\begin{table}[!ht]
    \centering
    \caption{Comparison of APE between cuSFM with and without Extrinsic Refinement on KITTI sequences}
    \label{tab:extrinsic_refinement}
    \begin{tabular}{lccccccc}
    \toprule
    \multirow{2}{*}{Sequence} & \multirow{2}{*}{Method} & \multicolumn{6}{c}{Absolute Trajectory Error (m)} \\
    \cmidrule(lr){3-8}
    & & RMSE & Mean & Median & STD & Min & Max \\
    \midrule
     \multirow{2}{*}{00} 
    & CuSFM w.o. ER & 1.388 & 1.189 & 0.985 & 0.716 & 0.158 & 3.382 \\
    & CuSFM w/ ER & \textcolor{nvidia_green}{\textbf{1.066}} & \textcolor{nvidia_green}{\textbf{0.934}} & \textcolor{nvidia_green}{\textbf{0.793}} & \textcolor{nvidia_green}{\textbf{0.516}} & \textcolor{nvidia_green}{\textbf{0.118}} & \textcolor{nvidia_green}{\textbf{2.608}} \\
    \midrule
     \multirow{2}{*}{02}
    & CuSFM w.o. ER & 2.899 & 2.610 & 2.422 & \textcolor{nvidia_green}{\textbf{1.261}} & \textcolor{nvidia_green}{\textbf{0.419}} & 6.714 \\
    & CuSFM w/ ER & \textcolor{nvidia_green}{\textbf{2.821}} & \textcolor{nvidia_green}{\textbf{2.498}} & \textcolor{nvidia_green}{\textbf{2.189}} & 1.311 & 0.676 & \textcolor{nvidia_green}{\textbf{6.512}} \\
    \midrule
    \multirow{2}{*}{05}
    & CuSFM w.o. ER & 1.174 & 1.049 & 1.034 & 0.525 & 0.277 & 3.433 \\
    & CuSFM w/ ER   & \textcolor{nvidia_green}{\textbf{0.792}} & \textcolor{nvidia_green}{\textbf{0.751}} & \textcolor{nvidia_green}{\textbf{0.717}} & \textcolor{nvidia_green}{\textbf{0.249}} & \textcolor{nvidia_green}{\textbf{0.123}} & \textcolor{nvidia_green}{\textbf{1.407}} \\
    \midrule
    \multirow{2}{*}{06}
    & CuSFM w.o. ER  & 0.783 & 0.741 & 0.738 & 0.255 & 0.112 & 1.488 \\
    & CuSFM w/ ER   & \textcolor{nvidia_green}{\textbf{0.536}} & \textcolor{nvidia_green}{\textbf{0.488}} & \textcolor{nvidia_green}{\textbf{0.561}} & \textcolor{nvidia_green}{\textbf{0.220}} & \textcolor{nvidia_green}{\textbf{0.067}} & \textcolor{nvidia_green}{\textbf{0.789}} \\
    \midrule
    \multirow{2}{*}{09}
    & CuSFM w.o. ER & 2.040 & 1.725 & 1.268 & 1.089 & 0.395 &  3.882 \\
    & CuSFM w/ ER & \textcolor{nvidia_green}{\textbf{1.576}} & \textcolor{nvidia_green}{\textbf{1.407}} & \textcolor{nvidia_green}{\textbf{1.072}} & \textcolor{nvidia_green}{\textbf{0.711}} & \textcolor{nvidia_green}{\textbf{0.406}} & \textcolor{nvidia_green}{\textbf{3.133}} \\
    \bottomrule
    \end{tabular}
\end{table}

The experimental results demonstrate significant improvements across all evaluated sequences when extrinsic refinement is enabled. 
Notably, Sequence 05 shows the most substantial enhancement, with a 32.5\% reduction in RMSE (from 1.174m to 0.792m) and a 52.6\% improvement in maximum error (from 3.433m to 1.407m). 
Similarly, Sequence 06 exhibits a 31.5\% reduction in RMSE (from 0.783m to 0.536m) and a 47.0\% decrease in maximum error (from 1.488m to 0.789m). 
These improvements are particularly pronounced in sequences with complex trajectories, suggesting that extrinsic refinement is especially beneficial in challenging scenarios.

The results validate the effectiveness of our extrinsic refinement strategy and demonstrate that introducing extrinsic constraints between cameras can form rigid body constraints, thereby further enhancing trajectory accuracy. 
The consistent performance enhancement across all sequences, with improvements ranging from 2.7\% (Sequence 02) to 32.5\% (Sequence 05) in RMSE, demonstrates the robustness of our approach. 
The reduction in STD values, particularly notable in Sequence 00 (28.0\% reduction) and Sequence 05 (52.6\% reduction), indicates that extrinsic refinement not only improves overall accuracy but also enhances the consistency of trajectory estimates.

\subsubsection{Localization Mode}

In practical applications, mapping requirements exhibit significant diversity. While previous experiments demonstrated batch processing capabilities for single-session data collection, real-world scenarios often involve sequential data acquisition, necessitating the integration of newly collected data into existing maps. This functionality, commonly referred to as crowdsourced mapping in industrial applications, represents a fundamental capability for scalable mapping systems. Additionally, localization within pre-existing maps represents another critical requirement, particularly prevalent in robotic applications. CuSfM addresses these diverse requirements through its localization mode, which supports both map updating and frame-based localization capabilities. 

To validate these capabilities, experiments were conducted using SDG dataset. The dataset comprises three cameras mounted on an autonomous vehicle platform: one forward-facing camera and two rear-facing cameras on both sides. The data collection involves two distinct trajectories traversing the same intersection from different directions. This experimental setup demonstrates cuSfM's capability to support diverse camera configurations. Urban intersection scenarios represent a typical use case requiring multi-trajectory fusion, as they often involve complex spatial relationships and overlapping viewpoints from different approaches. 

\begin{figure}[!ht]
    \centering
    \includegraphics[width=\linewidth]{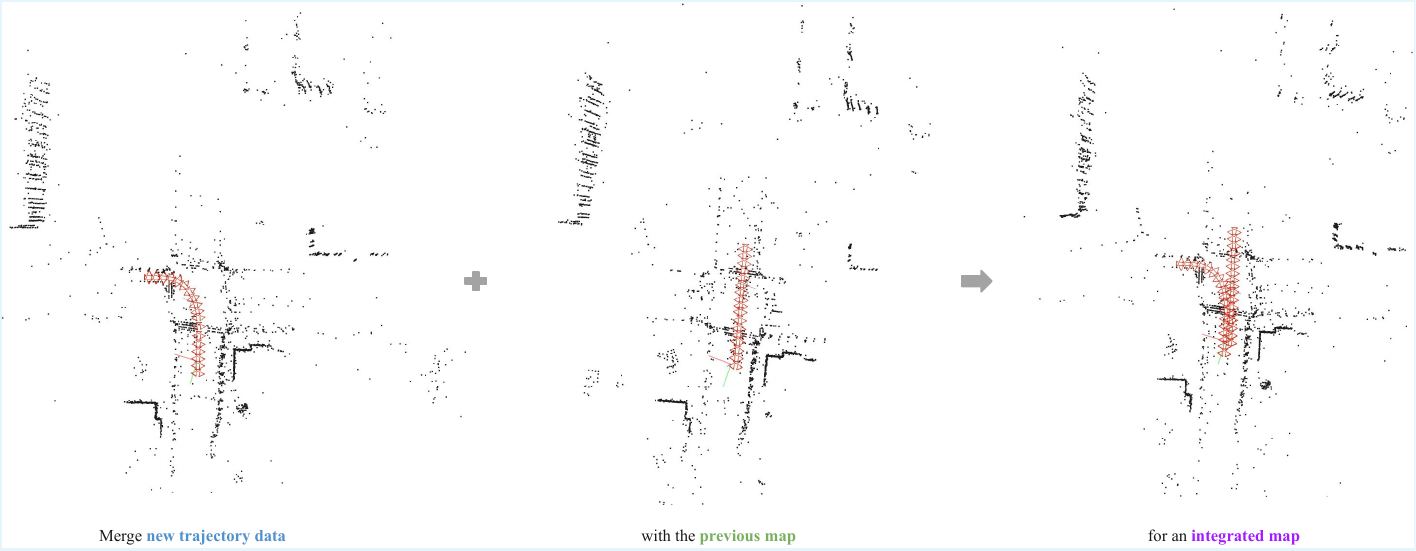}
    \caption{Illustration of map integration capabilities in cuSfM. Left: New trajectory data. Middle: Previous map. Right: Integrated map showing combined trajectory and updated environmental structure.}
    \label{fig:patch}
\end{figure}

Figure~\ref{fig:patch} illustrates these capabilities. The left panel displays new trajectory data, while the middle panel shows the previous map. Despite the short trajectory length used for demonstration purposes, the system supports significantly larger-scale maps. The spatial overlap between the new trajectory and the previous map indicates the feasibility of integrating new data into the existing map. The right panel demonstrates the integrated map, where the combined trajectory incorporates both the new trajectory data and the previous map information, while the environmental structure is updated to incorporate complementary information from different viewing angles.

CuSfM offers two operational modes for existing maps: adjust mode and fixed mode, in addition to supporting single-frame localization against the map. To the best of our knowledge, no existing open-source SfM project provides such comprehensive functionality for diverse mapping and localization scenarios.

\section{Conclusion}
This paper presents cuSfM, a CUDA-accelerated Structure-from-Motion system that addresses the computational efficiency and accuracy challenges in visual mapping and localization. The system introduces a non-redundant data association strategy that leverages pose graph priors to construct optimal view graphs, while incorporating a novel stereo relative pose estimation algorithm that directly estimates translation scale using only 2D observations. The framework provides comprehensive support for diverse operational modes including map updating, frame-based localization, and extrinsic parameter refinement through GPU-accelerated feature extraction and matching combined with efficient bundle adjustment optimization. Extensive evaluations demonstrate the system's effectiveness across multiple camera configurations and crowdsourced mapping scenarios on both real-world and simulated datasets. The open-source implementation enables practical deployment for visual mapping and localization applications, with future development focusing on dynamic environments and large-scale urban mapping scenarios.

\clearpage
\setcitestyle{numbers}
\bibliographystyle{plainnat}
\bibliography{main}

\end{document}